# Intent Matching based Customer Services Chatbot with Natural Language Understanding

Alvin Chaidrata
*School of Computer Science*
*University of Nottingham Malaysia*
Semenyih, Malaysia
hfyac1@nottingham.edu.my

Mariyam Imtha Shafeeu
*School of Computer Science*
*University of Nottingham Malysia*
Semenyih, Malaysia
hfyms2@nottingham.edu.my

Sze Ker Chew
*School of Computer Science*
*University of Nottingham Malaysia*
Semenyih, Malaysia
kczskc@nottingham.edu.my

Zhiyuan Chen
*School of Computer Science*
*University of Nottingham Malysia*
Semenyih, Malaysia
zhiyuan.chen@nottingham.edu.my

Jin Sheng Cham
*School of Computer Science*
*University of Nottingham Malaysia*
Semenyih, Malaysia
hfyjc7@nottingham.edu.my

Zi Li Yong
*School of Computer Science*
*University of Nottingham Malysia*
Semenyih, Malaysia
hfyzy2@nottingham.edu.my

Uen Hsieh Yap
*School of Computer Science*
*University of Nottingham Malaysia*
Semenyih, Malaysia
hcyuy1@nottingham.edu.my

Dania Imanina Binti Kamarul Bahrin
*School of Computer Science*
*University of Nottingham Malysia*
Semenyih, Malaysia
efydk1@nottingham.edu.my

*Abstract*— Customer service is the lifeblood of any business. Excellent customer service not only generates return business but also creates new customers. Looking at the demanding market to provide a 24/7 service to customers, many organisations are increasingly engaged in popular social media and text messaging platforms such as WhatsApp and Facebook Messenger in providing a 24/7 service to customers in the current demanding market. In this paper, we present an intent matching based customer services chatbot (IMCSC), which is capable of replacing the customer service work of sales personnel, whilst interacting in a more natural and human-like manner through the employment of Natural Language Understanding (NLU). The bot is able to answer the most common frequently asked questions and we have also integrated features for the processing and exporting of customer orders to a Google Sheet.

I. INTRODUCTION

Currently, the majority of customer interaction is via WhatsApp texts and calls with sales personnel manually handling enquiries and entering the orders into their system. This leaves room for human error and delays, as sales personnel are not infallible and can occasionally be too busy to deal with customers in a timely manner. Furthermore, at present, there is no way of answering customer enquiries and taking orders outside of working hours. Hence, our project aims to solve these issues by creating a human-centric customer service chatbot to serve customers better and improve business efficiency.

The purpose of this project is to develop an intent matching based customer services chatbot (IMCSC), which can facilitate and handle the daily routine work of sales personnel. This includes:

- Creating and integrating the chatbot with WhatsApp, Facebook Messenger, and Telephony using Dialogflow
- Employing natural language understanding (NLU) to manage and respond to customer inputs in a human-like manner
- Processing customer orders and recording customer orders into Google Sheets for the client.

II. LITERATURE REVIEW

According to [1], it is estimated that 1.4 billion people currently use social media messaging apps and are willing to use chatbots. Furthermore, as reported in [2] 80% of the businesses surveyed already employed chatbots or planned to employ them by the year 2020. With applications ranging anywhere from e-commerce, education, health, and even finance; this growth can be seen by the fact that the global chatbot market is expected to reach USD 1.25 billion by 2025.

One of the main complaints regarding chatbots is that they are too "unnatural", which has led to companies investing in Natural Language Processing (NLP) [3] and Natural Language Understanding (NLU) [4] in order to provide consumers with a more human- like conversational chatbot experience. As discussed in [5], it costs businesses USD 1.3 trillion to service the 265 billion customer service requests made every year, and the use of chatbots can help these companies save up to 30% in customer service- related costs.

There are a few current implementations of chatbots and other software that can help to build chatbots. Some companies claimed to use Dialogflow [6] to help businesses in optimizing their customer support workflows by automating responses to repeat inquiries. Livechat has created a software called BotEngine, which can help eCommerce businesses answer customer support inquiries that are unique to their businesses using natural language processing (NLP) and an interface that gives users the ability to build their own chatbot. This is a user-built chatbot software, just like Dialogflow. An example of a chatbot that uses NLP to create a conversational experience is Bus Uncle [7]. The Bus Uncle chatbot fills a niche as it allows users to query bus schedules in a unique and human-like interaction. This shows that chatbots can revolutionize user experience by transforming a simple search into an interaction with a human-like agent.

## III. IMCSC OVERVIEW

### A. Product Perspective

The IMCSC is in the form of the Dialogflow agent which is trained using machine learning techniques and integrated with webhook script to handle backend logics. The frontend user interface (UI) components are the messenger platforms that we integrate with the agent such as WhatsApp, Facebook Messenger etc.

The user can type and interact with the agent via the messenger app UI. When an end-user writes or says something, Dialogflow matches the end-user expression to the best intent in the agent. Matching an intent is also known as intent classification. Intent matching is based on the Dialogflow natural language understanding model that is used to train the agent. When an intent is matched at runtime, Dialogflow provides the extracted values from the end-user expression as parameters. Each parameter has a type, called the entity type, which dictates exactly how the data is extracted. The extracted data can then be used to perform logical operations or generate responses. The response is then sent back to the messenger app to be displayed to the end-user through the user interface.

### B. Product Functions

The main purpose of the IMCSC is to ease the workload for the customer service employees. With the IMCSC, potential customers can obtain answers regarding product delivery, payments, as well as general information regarding the business such as store location and contacts. Aside from that, customers can also make orders through the IMCSC. Lastly, it can also answer the users in a human-like manner as well as engage in simple small talks.

### C. User Characteristics

The user type that will be interacting with the chatbot are WhatsApp and Facebook Messenger users. These users are assumed to be familiar with these messaging services and will be able to interact with the IMCSC in order to ask enquiries or place orders.

### D. Assumptions and Dependencies

The user is assumed to have a stable internet connection when using the system. Another assumption made is that the user is already familiar with the products offered by the supplier and are recurring customers. The user should have basic understanding on how messaging apps such as WhatsApp and Facebook Messenger work.

It is also assumed that the product's data will be made available for the project in some phase of its completion in the future. Until then, test data created by us will be used for providing the demo for presentations. Other than that, it will be assumed that users are familiar with using an internet browser.

## IV. IMCSC DESIGN SPECIFICATIONS

In this section, the requirement specifications, software quality attributes and use case diagram of IMCSC will be explained.

### A. Requirement specifications

*1) Functional Requirement-Mandatory (FRM)*

*a) FRM1:* Users should be able to converse with the chatbot (SD1,SD2,SD3,SD4)

*b) FRM2:* Chatbot should be able to converse with the user in a human-like manner (many variations of responses for a given query) (SD1,SD2,SD3,SD4)

*c) FRM3:* Users should be able to search for products via the chatbot (SD1)

*d) FRM4:* Users should be able to obtain relevant business info (Operating hours, Contact number, Location of shop) via the chatbot (SD1)

*e) FRM5:* Users should be able to obtain answers for questions listed in the business' FAQ via the chatbot (SD1)

*f) FRM6:* Chatbot should be able to communicate in English (SD1,SD2,SD3,SD4)

*2) Functional Requirement-Desireable (FRD)*

*a) FRD1:* Users should be able to converse with the chatbot (SD1,SD2,SD3,SD4)

*b) FRD2:* Chatbot should be able to converse with the user in a human-like manner (many variations of responses for a given query) (SD1,SD2,SD3,SD4)

*c) FRD3:* Chatbot should be able to communicate in languages other than English (Chinese)

*d) FRD4:* Chatbot should be able to identify repetitive questions from the customer

*3) Functional Requirement-Optional (FRO)*

*a) FRO1:* Chatbot should be able to detect non-supported language from the customer and advise the customer to use only supported language.

*b) FRO2:* Chatbot should be able to end the conversation if spam is detected

*4) Non-Functional Requirements (NFR)*

*a) NFR1:* Chatbot should be built using DialogFlow as the foundation

*b) NFR2:* Chatbot should be able to integrate with Facebook Messenger, WhatsApp, and telephony system

*c) NF3:* Chatbot should be able to utilize text-to-speech function in order to integrate with the telephony system

*d) NFR4:* Chatbot should be available 24/7, except when interrupted by unexpected errors (server failure, runtime failure, random errors)

B. *Software quality attributes*

- Correctness: The Chatbot should never allow anyone to read messages not intended for that person.
- Usability: The bot should be able to understand the user as much as possible, allowing them to accomplish their goals without getting error prompt.
- Functionality: The bot should meet the mandatory requirements of the requirement specification.
- Extensibility: The bot should allow for easy addition or expansion of current functionalities without having to make major changes to the existing bot.
- Maintainability: The bot should be structured in a way that can be easily understood by a new developer. Its functionality should also be grouped or categorized in such a way that makes it easy to track and fix errors.

C. *Use case diagram*

The IMCSC use case diagram has been shown in Fig 1.

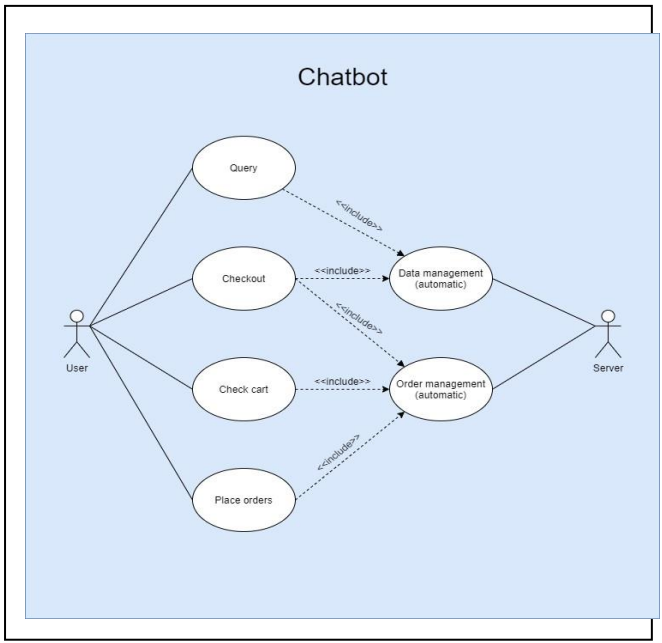

Figure 1. IMCSC use case diagram.

## V. IMCSC IMPLEMENTATION

The IMCSC is constructed using the Dialogflow platform as the created agents can be easily trained using Google Cloud's Natural Language API. Additionally, Dialogflow allow us to easily integrate the chatbot with popular messaging platforms. The IMCSC system architecture has been show in Fig 2.

In Dialogflow, NLU modules called agents, that are responsible for handling conversations with end users. The platform translates end-user text or audio during a conversation to structured data that your apps and services can understand. Dialogflow relies on two concepts to perform the NLU operations required: intents and entities [8]. Entities are tools used for extracting parameter values from NLU inputs whereas intents represent the mapping between the input from a user and the action that should be taken by the chatbot. Each agent can have various defined intents to which Dialogflow can then match the input given. There are many techniques can do intent matching, such as using techniques in [9]. In this project, the process of matching these intents is referred as intent classification.

Training phrases are then added to facilitate intent classification. For example, the training phrase "when do you close?" helps the agent recognise similar input such as "closing hours" or "what time will you close". Coming up with every possible input expression is not required with Dialogflow as the built-in machine learning adds to the list of phrases in order to provide better intent matching. Although some entity types such as 'product category' have to be created manually, Dialogflow also annotates parts of these training phrases to assign entity types such as time, date, place etc. which the agent uses to provide various answers depending on the entity type. For instance, Dialogflow matches the words 'tomorrow' and 'now' in the training phrases "closing time tomorrow" and "are you open now" to the date and time entity types respectively.

Furthermore, Dialogflow allows the use of fuzzy search which is implemented by adding synonyms to the different entities. This ensures that the IMCSC is able to recognise product names even if the customer has used a slightly different name.

A. *Intent*

Intents are used in the IMCSC to return the appropriate response based on the user's query. Product Info sub-agent handles product specific queries and functions such as order taking, product and category checking. Following are the intents for our Product Info sub-agent.

- Find-Products - To find the product names from the database given by the client. Users are required to search the product according to its brand and category.
- orderTaking - To obtain the product name and quantity of the item from the user to place an order.
- itemConfirm - To confirm the item ordered by the user and store the item details (product name and quantity) into the cart.

- itemDecline - To handle the flow of the conversation if the user declines the item when prompted to confirm.
- cartChecking - To let the user see what items they have ordered.
- orderCancel - To remove a specific item from the cart. The item name should be given by the user.
- Personal-Info - To obtain the name, shipping address, and phone number of the user during checkout. All fields are mandatory. The personal details will be sent to the database alongside the items in the cart.

FAQ sub-agent is responsible for answering queries about the business with answers that can be found in the FAQ section of the business website. Business Contact sub-agent handles queries regarding the business operations.

### B. Entities

Entities are used to link many terms to one word, so that the IMCSC can recognize synonyms of the user's query, while ensuring that the chatbot only accepts terms that are in entities.

Following are the entities for the Product Info sub-agent:
- ProductName - This entity consists of all of the product names available in our client's shop.
- Brand - This entity consists of all of the brands available in our client's shop
- Category - This entity consists of all of the category available in our client's shop
- Yes - This entity consists of all of the phrases that show agreement on something.
- No - This entity consists of all of the phrases that show objection to something.

### C. Facebook Messenger Integration

This integration is done by creating a Facebook app that uses the Facebook Messenger Platform. The Facebook app has been configured to communicate with the mega agent on DialogFlow. Hence, when an individual sends a message on Messenger, the Facebook server will make a webhook call to the Facebook messaging app hosted on the business server which passes the message to DialogFlow agent. Then, the mega agent's response is sent through the Facebook messaging app and to the user on Messenger using the Facebook Messenger API. In this manner, we are able to build guided conversations for the users to serve as a bridge between our agent and the client's business presence on Messenger.

### D. WhatsApp Integration

As Dialogflow does not support direct integration with WhatsApp, we have made the connection through an open source, third-party API. According to our research, Twilio is one of the most preferred and common third-party API to integrate with DialogFlow for a WhatsApp chatbot. Twilio API allows software developers to programmatically make and receive phone calls, send and receive text messages, and perform other communication functions. DialogFlow has built in Twilio integration and Twilio supports WhatsApp integration through its API.

### E. Testing

TABLE I. SAMPLE FUNCTIONAL REQUIREMENTS TESTING RESULTS

| Requirement | Description | Expected Results | Pass/Fail |
|---|---|---|---|
| Inquiries | Bot answers common user's question such as business hours, items, policies e.t.c | Bot is able to answer the user's question or responds with a default message if not answerable | Pass |
| Ordering | Bot allows user to initiate order | Bot allows user to make and cancel orders | Pass |
| Identifying Unavailable Products | Bot is able to identify if an item isn't in the business product's database | Bot tells the user that the product desired isn't available | Pass |
| Fuzzy Matching Product and Brand Names | Bot is able to identify if a user misspells a product/brand name during order | Bot recognizes the product and saves it into the user's cart | Pass |
| Handling unknown intent | Bot handles unknown intent by responding with a default prompt | Bot responds with a default prompt requesting the user to rephrase their message | Pass |

The biggest advantage chatbots have over humans is that they do not exhibit behavioural errors like frustration, bias or judgment. However, the biggest downside is that chatbots cannot accurately assess the situational cues of a conversation. Hence, during testing and designing of IMCSC, we apply the Pareto Principle (80:20 rule). The main idea is that the chatbot should respond well to the 20% of the highly repetitive use cases that are more than 80% of the volume of use cases [10]. For the rest of the use cases, we cover it with the default fallback intent which responds to all the improper user inputs.

Therefore, for testing, the IMCSC is tested for the following:
- Chatbot should respond to basic salutations and introductions.
- Chatbot should respond to common variations of keywords used in intents.
- Chatbot should not be stuck in a cyclic loop if a condition fails repeatedly. This could happen during the product order loop.

- The chatbot should be able to handle basic typos and still match the user input to the correct intent.
- The chatbot should be able to comprehend basic natural language for single or multiple entities.

For each intent, multiple test cases were written for each intent so that we can test each scenario that could occur. Entities are used to link many terms to one word, so that the chatbot can recognize synonyms of the user's query, while ensuring that the chatbot only accepts terms that are in entities. The sample functional requirements testing results have been show in Table 1.

## VI. CONCLUSION

All mandatory requirements and especially these integrated features through WhatsApp and Facebook Messenger have been successfully delivered. To conclude, the IMCSC is able to converse with customers in a conversational manner instead of the usual menu driven interface. It also can answer customer enquiries regarding refunds, deliveries, payment, location, business hours, and some other common business-related questions. Furthermore, it can even provide customers with a Waze link to the shop location or inform them whether the store is currently open or close.

On the product side, customers can ask the IMCSC if the business offers the item they are looking for. The chatbot is also able to record single or multiple orders at once into a Google Sheet. When customers are ordering, it can check what they have ordered and cancel it if necessary. Once customers are sure of their orders, the IMCSC will take their personal information and delivery details which will then be recorded into the Google Sheet. Customers are also able to converse with the IMCSC through messaging platforms either via WhatsApp or Facebook Messenger. Lastly the testing results justified the completeness and robustness of the IMCSC.

## ACKNOWLEDGMENT

We would like to thank our industry collaborator, Shing Tik Vegetarian Sdn. Bhd, for the support to the whole project. This paper is sponsored by the School of Computer Science Staff Conference Fund from the University of Nottingham Malaysia.

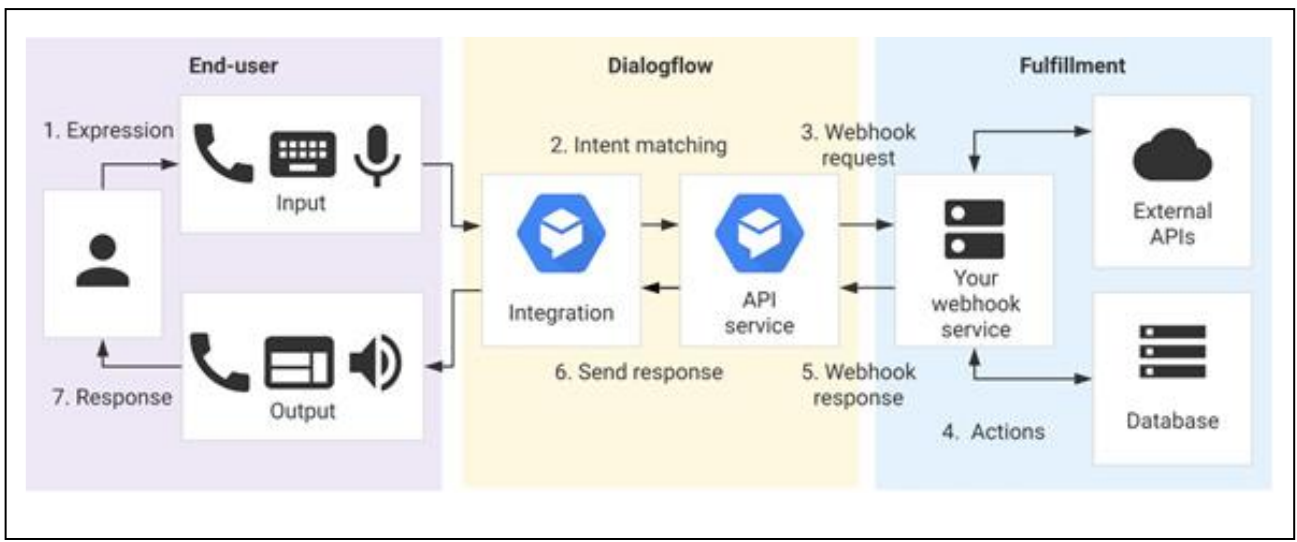

Figure 2. IMCSC system architecture.